\documentclass{article}

\usepackage{arxiv}

\usepackage[utf8]{inputenc} 
\usepackage[T1]{fontenc}    
\usepackage{hyperref}       
\usepackage{url}            
\usepackage{booktabs}       
\usepackage{amsfonts}       
\usepackage{nicefrac}       
\usepackage{microtype}      
\usepackage{lipsum}
\usepackage{graphicx}
\usepackage{subfig}
\usepackage{textcomp}
 
\newcommand{\iva}{IVA}

\title{Detecting Alzheimer{\textquotesingle}s Disease by estimating attention and elicitation path through the alignment of spoken picture descriptions with the picture prompt}

\author{ 
    Bahman Mirheidari\\
  Department of Computer Science\\
  University of Sheffield\\
  United Kingdom \\
  \texttt{b.mirheidari@sheffield.ac.uk} \\ 
   \And
   Yilin Pan \\
  Department of Computer Science\\
  University of Sheffield\\
  United Kingdom \\
  \texttt{yilin.pan@sheffield.ac.uk} \\ 
   \And
 Traci Walker \\
  Department of Human Communication Sciences\\
  University of Sheffield\\
  UK \\
  \texttt{traci.walker@sheffield.ac.uk} \\
   \And
 Markus Reuber \\
  Academic Neurology Unit, Royal Hallamshire Hospital\\
  University of Sheffield\\
  United Kingdom \\
  \texttt{m.reuber@sheffield.ac.uk} \\
  \And
 Annalena Venneri \\
  Sheffield Institute for Translational Neuroscience\\
  University of Sheffield\\
  United Kingdom \\
  \texttt{a.Venneri@sheffield.ac.uk} \\
  \And
 Daniel Blackburn, Annalena Venneri \\
  Sheffield Institute for Translational Neuroscience\\
  University of Sheffield\\
  United Kingdom \\
  \texttt{d.blackburn@sheffield.ac.uk} \\
   \And
   Heidi Christensen \\
  Department of Computer Science\\
  University of Sheffield\\
  United Kingdom \\
  \texttt{heidi.christensen@sheffield.ac.uk}  
}

\begin{document}
\maketitle

\begin{abstract}
Cognitive decline is a sign of Alzheimer{\textquotesingle}s disease (AD), and there is evidence that tracking a person{\textquotesingle}s eye movement, using eye tracking devices, can be used for the automatic identification of early signs of cognitive decline. However, such devices are expensive and may not be easy-to-use for people with cognitive problems. In this paper, we present a new way of capturing similar visual features, by using the speech of people describing the Cookie Theft picture - a common cognitive testing task - to identify regions in the picture prompt that will have caught the speaker{\textquotesingle}s attention and elicited their speech. After aligning the automatically recognised words with different regions of the picture prompt, we extract information inspired by eye tracking metrics such as coordinates of the area of interests (AOI)s, time spent in AOI, time to reach the AOI, and the number of AOI visits. Using the DementiaBank dataset we train a binary classifier (AD vs. healthy control) using 10-fold cross-validation and achieve an 80\% F1-score using the timing information from the forced alignments of the automatic speech recogniser (ASR); this achieved around 72\% using the timing information from the ASR outputs.
\end{abstract}


\section{Introduction} 
The number of people living with dementia has increased significantly in recent years and the economic impact on the society is huge. According to \cite{patterson2018world} every 3 seconds a person develops dementia in the world. There are around 50 million people living with dementia in total and it is estimated to rise to 152 million by 2050; the current cost of dementia is approximated about a trillion US dollar a year.  

The treatments are most effective in the early stage of the disease before dementia has developed . However, the process of diagnosing this disorder is very complex, mostly due to overlapping symptoms with normal ageing and low accuracy of existing cognitive screening tools. The currently available tests for stratifying (screening) people with cognitive complaints, are based on pen-and-paper testing and lack sensitivity or specificity especially early in the disease process. It is therefore highly desirable to build a cheap and reliable automatic screening tool to identify people at risk of developing dementia. Ideally, such a tool could be used in a person{\textquotesingle}s own home, without the need for an examiner to be present.  

Speech and language are affected early in dementia (e.g. by losing vocabulary, simplifying syntax/semantics, overusing semantically empty \cite{Appell1982,Bayles1987,Hamilton1994}). The Boston Cookie Theft picture description task \cite{goodglass1972assessment} is a widely used cognitive test which was originally designed to capture decline in spontaneous speech and language of people with aphasia. However, it has been shown that it could help in identify people with dementia as well \cite{giles1996performance, forbes2005detecting}. In this task, the subjects are asked to look at a picture (the \textsl{prompt}) and describe everything happening in the picture. 

These tests are then normally scored based on the utterances that are generated in the description, and by counting things like the number of simple/complex and complete/incomplete sentences. However, there is no part of the assessment scoring that takes into account the picture prompt itself, i.e., things like which areas of the picture the person might have been looking at, in which order. However, it is known that cognitive decline may affect a person's eye movement and attention \cite{lai2013review}. Since the generated language is a direct result of the person looking at the the picture prompt, tracking the attention and elicitation path on the picture might reveal informative data to help identify early signs of dementia. Accurate tracking of eye movement requires an eye tracking devices. However, routinely using eye tracking devices in memory clinics or for home-based testing would be an impractical, expensive and inconvenient solution. This paper proposes an approach to capturing information about eye movement path on a picture prompt without needing eye tracking equipment.


The approach uses the automatic transcription of the picture description (using automatic speech recognition (ASR)) to identify the area of interests (AOIs) in the picture prompt, that may have caught the attention of the individual and elicited their speech at  that particular point in time. A number of features will be used to support the estimation. The extracted features are then used to train a classifier to identify dementia and compared with using features designed to capture information directly from the speech itself. 

In the remainder of the paper, Section~\ref{sec:background} presents the background, Section~\ref{sec:tracking_attention} presents our proposed approach, Section~\ref{sec:expsetup} describes the experimental setup, and finally, results and conclusions are presented in Sections~\ref{sec:results} and \ref{sec:conclusions}.   
\section{Background}  
\label{sec:background}  
Eye tracking technology has been used in a variety of applications ranging from learning assistants \cite{lai2013review,van2013eye}, Human Computer Interface (HCI) design \cite{chandra2015eye,lopez2015eye}, mobile phone applications \cite{krafka2016eye}, assistive technology for disabled people \cite{raudonis2009discrete}, and diagnosing mental/memory diseases (autism, Mild Cognitive Decline (MCI), Parkinson{\textquotesingle}s and Alzheimer{\textquotesingle}s disease (AD), etc.)\cite{harezlak2018application}.  

According to a survey by \cite{beltran2018computational}, eye movement difficulties might be a sign of cognitive decline and altered eye movements indicate visual-spatial and executive function problems, although same behaviour can be seen in aged subjects as well. Eye tracking can be done in a scenario based (e.g. reading) or exploration task or a more naturalistic situation (e.g identifying objects in daily life). In scenario based tasks, the AD participants exhibited longer fixation duration than the Healthy Controls (HCs), and in naturalistic tasks, the AD patients had fewer overall fixations as well as a lower
proportion of relevant fixations. Based on a study by \cite{levy2018prosaccade} eye tracking devices could assess cognitive functions of people with AD. The AD subjects showed a longer pro-saccade\footnote{in a pro-saccade task, eyes first focus on a dot in the centre of a screen and then turn the gaze to a target stimulus as it appears.} latency and more anti-saccade\footnote{in an anti-saccade task, eyes first focus on a dot in the centre of a screen, but they will be asked to turn their gaze in the opposite direction of a stimulus.} error rates compared to the healthy controls (HC). In a longitudinal eye tracking study on AD subjects vs. HCs, \cite{crawford2015disengagement} found out that the AD individuals had a slower saccade reaction time compared to the HC group, however, over a 12 month period, the reaction time for both groups did not deteriorate. 
  
The `Cookie Theft' picture description task was originally part of the Boston Diagnostic Aphasia Examination \cite{goodglass1983boston} in which the examiner shows the Cookie Theft line drawing  to individuals
asking them to describe everything that is happening in the picture.
The Pitt Corpus - DementiaBank \cite{becker1994natural} is a collection of audio recordings and transcripts of 104 HC, 208 AD and 85 other diagnosis participants describing the Cookie Theft picture \footnote{the latest figures taken from their website:\\ https://dementia.talkbank.org/access/English/Pitt.html}. The study was carried out between 1983 and 1988 and many of the participants had several repeated sessions during the study, providing a longitudinal corpus. A few years after gathering the data, the diagnoses of the subjects were reviewed again and the final decisions were made by the specialist neurologists. The corpus is free to download from their website and use for research purposes. In recent years, there has been a number of studies based on it, mostly aimed at detecting dementia using the audio, text and language processing technologies.  
 
In \cite{Yancheva2015}, the corpus was used to extract a wide range of features (over 477 lexico-syntactic, acoustic, and semantic) and train an AD/HC classifier. They achieved an 81\% classification accuracy using all features and 92\% when they selected 40 informative features from the their feature set. \cite{yancheva2016vector} used GloVe word vectors (representing the meaning of words) to make 10 common clusters of the words (only nouns and verbs) in the training set of the dementia group as well as 10 common cluster of words in the HC (each cluster representing a topic or related words, e.g.~C0:~window, floor, curtains, plate, kitchen, D0:~cookie, cookies, cake, baking, apples).~Then, using the average scaled distance between the words of a given transcript from the test set and the created clusters, they extracted 20 semantic features.~In addition they calculated the `idea density' as the number of topics mentioned in the transcript divided by the number of words, and the `idea efficiency' as the number of topics mentioned in the transcript divided by the length of recording.~They gained 80\% accuracy and 80\% F1 score when they combined all their features with the lexicosyntactic and acoustic features from \cite{fraser2016detecting}.~\cite{AlHameed2016} achieved a range of accuracies between 83\% and 92\% for a number of different classifiers trained on 263 acoustic-only features extracted from the audio files of DementiaBank, thereby avoiding the need to use ASR. Using the 20 most informative features their best classifier (BayesNet) achieved 95\%.~In their follow-up research they included the MCI group for classification \cite{al2017detecting} and gained an accuracy rate between 89.2\% and 92.4\% for the pairwise classifications.~\cite{hernandez2018computer} chose 25 HC transcriptions to make reference Information Content Units (ICUs) (representing the main information in the picture including the actors, objects, actions and places e.g. woman/lady/mother/mommy make up a group named mother, and little/young-boy/kid make another group (boy). These reference ICUs were then used to estimate the coverage of informativeness of a description, i.e., how much did the main ICUs cover. They extracted the coverage and a number of linguistic features to train an SVM classifier. The binary classifier achieved an 81\% F1 score (excluding the 25 reference HCs and the subjects with MCI). Note that all of these studies used the manual transcriptions and did not used the ASRs.

In addition to the conventional classifiers, a number of authors have tried using Deep Neural Network (DNN) based classifiers to detect dementia from the DementiaBank corpus.
In \cite{mirheidari2018detecting} we used the predefined GloVe word vectors to make a sequence classification, combining Convolutional Neural Network (CNN) and Long Short Term Memory (LSTM). The classifier achieved a 75.6\% accuracy, however, replacing the manual utterances with the automatic transcripts produced by the ASR (45\% WER), the accuracy dropped to 62.3\%.~\cite{karlekar2018detecting} trained three different DNN models to classify dementia:~LSTM, 2 Dimensional CNN, and a mix of LSTM and CNN (CNN-LSTM). They used the manual transcriptions and the part of speech (POS) tags provided for the words in the utterances. Using only the words in the transcriptions, CNN-LSTM model could outperform the two other models with an 84.9\% accuracy. However, the best performance achieved a 91.1\% accuracy when they used the POS tagged information to trained the CNN-LSTM model. 

\section{Tracking attention on picture prompt}
\label{sec:tracking_attention}




Unlike approaches described in the previous section, this paper presents a way to make use of the output of the ASR to estimate what people might have been looking at (their AOIs or attention) as they are describing the Cookie Theft picture. First, a number of reference AOIs were identified manually on the picture, corresponding to the important actors, objects and actions (nouns and verbs), e.g., boy, girl, mother, cookie, grab, fall, and wash. As we divide data into train and test sets, this information should be filtered based on the train set data, i.e., the words not included in the train set were removed from the reference AOIs. Thus the filtered reference AOIs indicates the location and size of the AOI for a given word.
After training an ASR on the training set data, we can drive the timing information for each word in an utterance, i.e., the forced alignments give the estimated start and end times for the words in the training set, and the output of the ASR (decoding latices) provides the timing information for the test set words in a similar way.  

\setlength{\textfloatsep}{2mm}
\begin{figure}[ht] 
    \centering
    \subfloat{(a){\includegraphics[width=6.5cm,height=4cm]{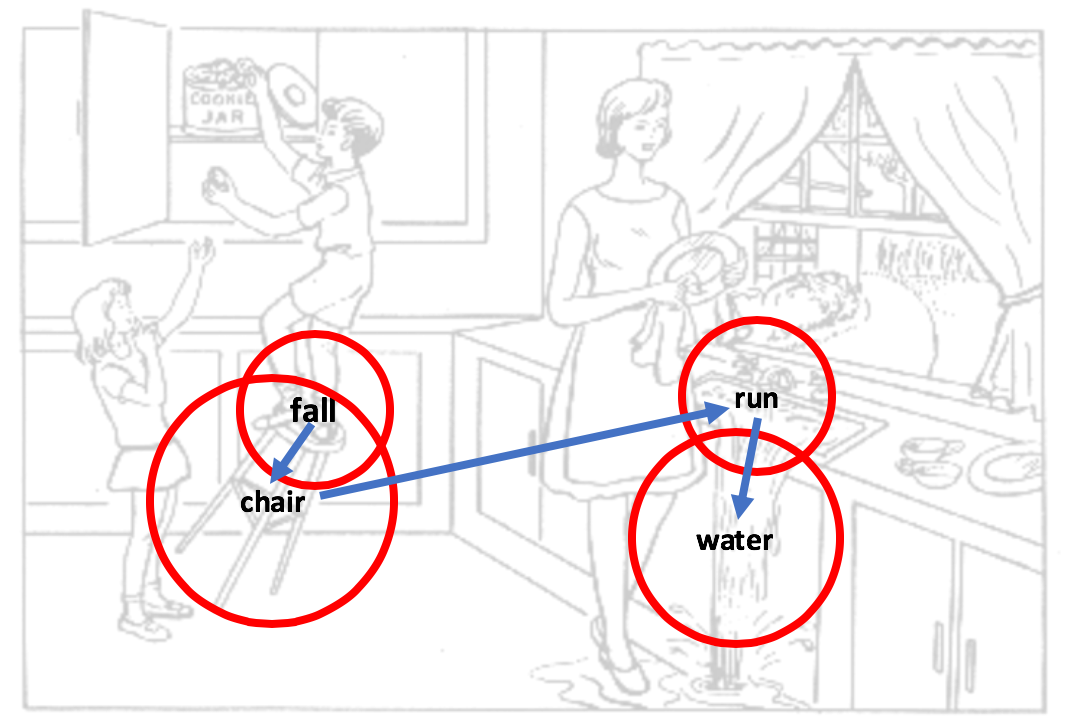}}}%
    \qquad
    \subfloat{(b) {\includegraphics[width=6.5cm,,height=4cm]{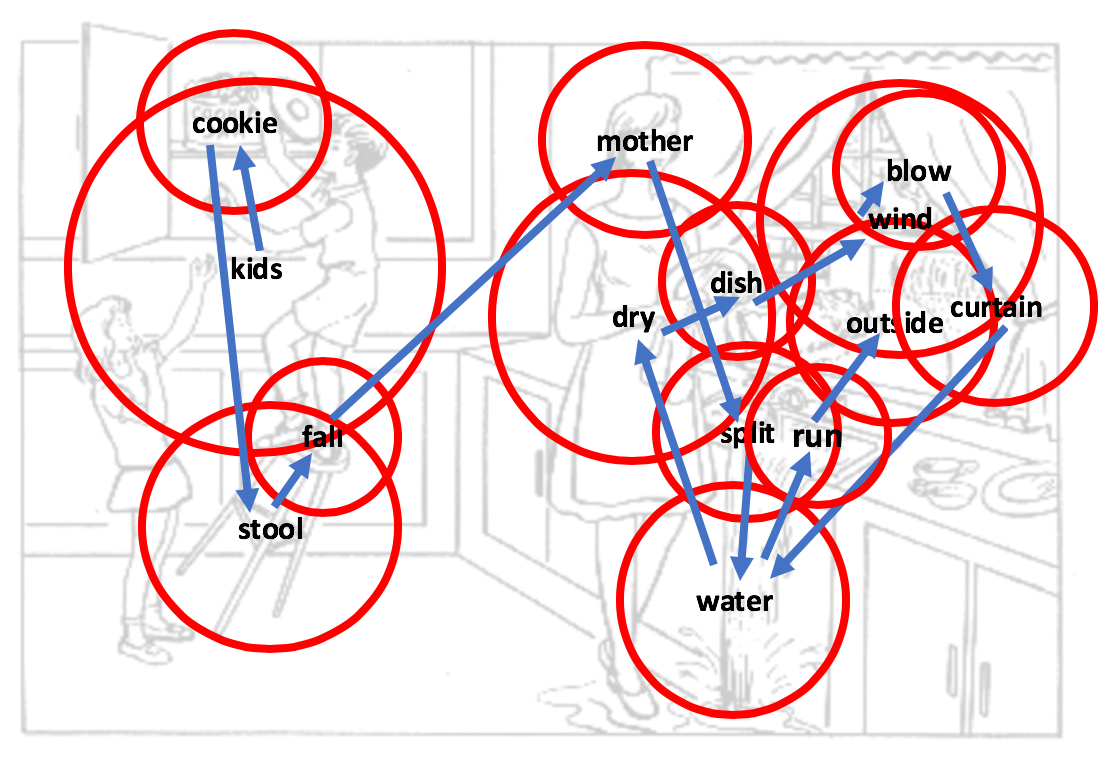} }}%
    \caption{Scanpath for the patient with AD 005-2 (a) vs. the health control 631-0 (b).}
    \label{fig:pics}%
\end{figure}  


Eye tracking features are often related to \textsl{saccades}, \textsl{fixations} and \textsl{gaze points} where gaze points are the points on a screen that our eyes directly stare at, fixations are the clusters of continuous and close gaze points (the pauses of eyes movements) over AOI on a screen, and saccades are rapid eye movements between fixations or the transitions from one gaze point to another \cite{salvucci2000identifying,blascheck2014state}. Fixation can be represented by the coordinates (x,y) and duration (time spent) normally in milliseconds. Scanpath is a way to visualise attention tracking, in which fixation is represented by a circle with a radius of the duration of fixation and a saccade is shown by a line connecting two fixations \cite{blascheck2014state}. Figure \ref{fig:pics} shows the scanpaths for the descriptions given by an individual with AD ((a), participant 005-2) and a healthy control ((b), participant 631-0). Participant 005-2 said:~``OK, He{\textquotesingle}s \textbf{falling} off a \textbf{chair}. She{\textquotesingle}s uh \textbf{running} the \textbf{water} over. Can{\textquotesingle}t see anything else. No, OK, She{\textquotesingle}s No.''. The important words (in bold) are identified by circles on the picture and the arrows show the transition from one AOI to another, representing the order of attention as well. Therefore, looking at the picture we can estimate which areas of the pictures were viewed by the participant over time. The utterance of the healthy control 631-0 is:``The \textbf{kids} are in the \textbf{cookies}. The \textbf{stool} is \textbf{falling} over, The \textbf{mother}{\textquotesingle}s \textbf{spilling} the \textbf{water} and also \textbf{drying} the \textbf{dishes} and the \textbf{wind} might be \textbf{blowing} the \textbf{curtains}. And, the \textbf{water}{\textquotesingle}s \textbf{running}. Uh I can{\textquotesingle}t tell that{\textquotesingle}s tell is anything going on \textbf{outside} or not. I guess that's all I see that{\textquotesingle}s not very many''. As can be seen, from the estimated scanpaths, the HC participant gave more details of the picture, visiting more reference AOIs (note that the AOI ``Water'' was visited twice by the participant).

\begin{figure*}
    \centering
    \subfloat{(a){\includegraphics[width=3.8cm]{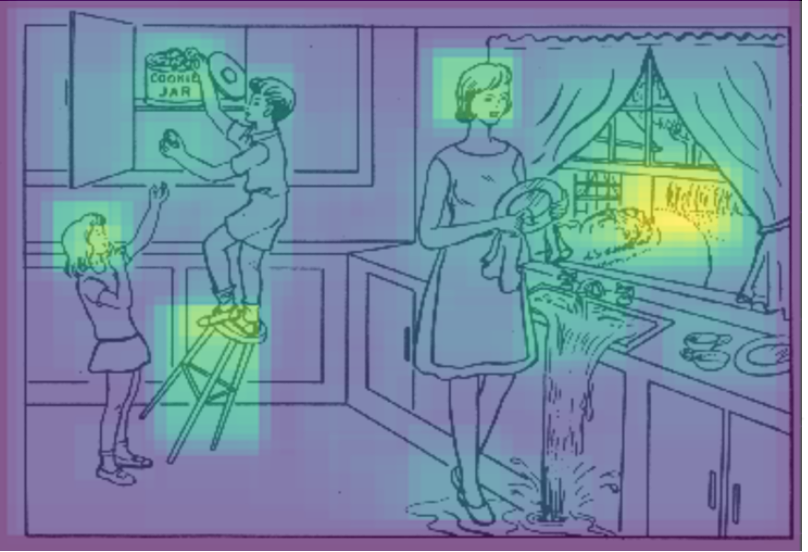} }}%
    \subfloat{(b) {\includegraphics[width=3.8cm]{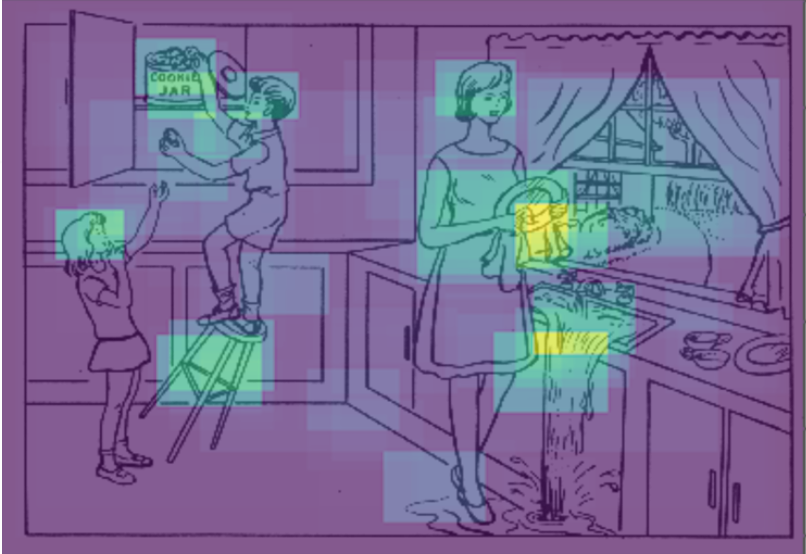}}}%
    \subfloat{(c) {\includegraphics[width=3.8cm]{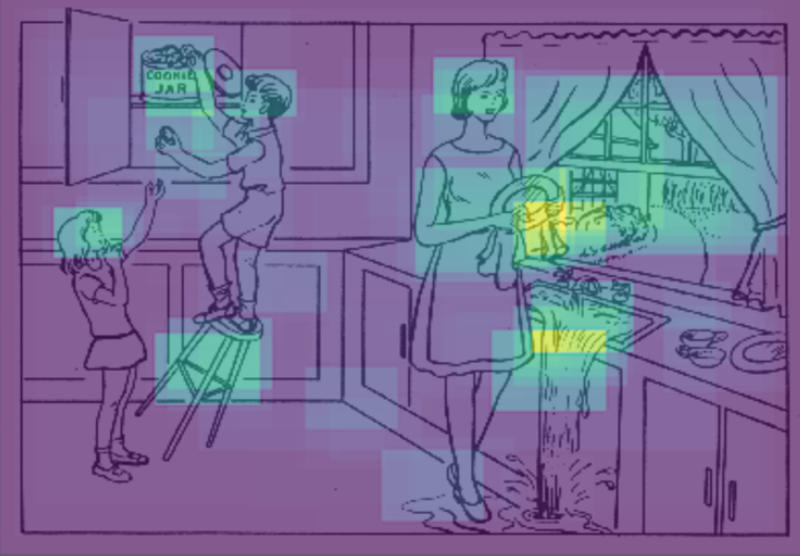} }}%
    \subfloat{(d) {\includegraphics[width=4.03cm]{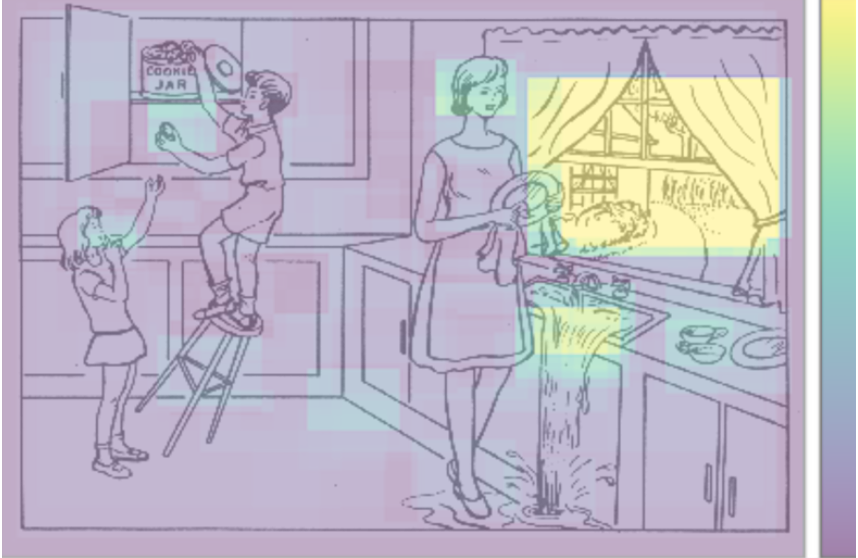}}}%
    \caption{Heatmaps of the area of interests (AOIs) (a) manually assinged as the references, (b) produced automatically for the AD group, (c) produced automatically for the HC group,  the difference between the AD and HC groups.}
    \label{fig:heatmaps}%
\end{figure*}

We hypothesise that the people with cognitive decline might not able to identify some of the important AOIs in the picture. Heatmaps \footnote{two dimensional graphical representation of data using colours, normally the warmer the colour (e.g. yellow warmer than blue) the higher the value.} represent the viewing behaviour and distribution of attention for the individuals watching a scene on a screen \cite{bojko2009informative}. 
 Figure \ref{fig:heatmaps} shows (a) the heatmaps of the AOIs which manually assigned AOIs on the picture, (b) the heatmaps produced automatically for the AD participants using the ASR and the reference AOIs, (c) the automatic heatmaps for the HC groups, and  the differences between the two groups. The heatmaps show that the AD groups, similarly to the HC group, could also identify many important AOIs, however, the differences are mostly on the part of the picture which shows a window and outside view including a path, trees and neighbours houses. Thus there could be some AOIs which were overlooked by the AD participants and did not catch their attention.

\section{Experimental setup} 
\label{sec:expsetup}
\textbf{Data:} DementiaBank contains 551 transcriptions/audio files from 98 HC (241 recordings) and 195 people with dementia (310 recordings) (including AD, other types of dementia and MCI). However, the diagnosis of a number of the participants were changed over the longitudinal study, i.e., some changed from MCI to AD, some from HC to MCI. For this evaluation, we have chosen to exclude those participants. This resulted in us using a total of 257 participants with stable diagnoses for the HC/AD classification experiments (89 HC participants - 215 recordings; 168 AD participants - 249 recordings)~\footnote{Note that \cite{unpublishedYilinPan} worked on the same dataset, but with 222 HC and 257 AD recordings.}. For training the ASR, all of the 551 recordings were used (diagnostic class is not relevant) in a 10-fold cross-validation approach. Additionally, we supplemented with two data sets recorded in-house: a collection of interviews with doctors (more than 64 hours of speech) \cite{mirheidari2018detecting}, and a much smaller collection of Cookie Theft description collected using an Intelligent Virtual Agent (\iva\ - 'digital doctor') \cite{mirheidari2017}. We refer to this data set as \iva{} and DementiaBank as Dem for the remaining parts of the paper.~Note that only 33 out of 76 \iva{} participants were diagnosed as AD or HC (17 vs. 16).
Table \ref{tab:datasets_info} gives more details of the data sets.  

\begin{table}[h]
\caption{\label{tab:datasets_info} Information about the datasets used for training the ASRs, including Len.:the total length in hours/mins, Utts.:number of utterances, Spks.:number of speakers, and Avg. Utts.:Average utterance length in seconds.}
\centering{
\begin{tabular}{p{2.1cm}  |p{1.1cm}  | p{0.4cm}|p{0.4cm}  | p{0.4cm}}
\specialrule{0.8pt}{0.8pt}{0.8pt} 
\multicolumn{1}{c|}{\textbf{Dataset(No)}} &  
\multicolumn{1}{c|}{\textbf{Len.}} & 
\multicolumn{1}{c|}{\textbf{Utts. }}&  
\multicolumn{1}{c|}{\textbf{Spks.}} & 
\multicolumn{1}{c}{\textbf{Avg Utts.}}\\ 
\specialrule{0.8pt}{0.8pt}{0.8pt}
\mbox{Dem (551)} &8h 34m&6737&293&4.6s \\ 
\hline
\mbox{Dr intvws (295)}&64h 21m&39184&736&5.9s\\
\hline
\mbox{\iva{} (76)}&1h 15m&497&76&9.11s\\  

\specialrule{0.8pt}{0.8pt}{0.8pt}
\end{tabular}
} 
\end{table}   

\textbf{Features:} For each AOI in the picture, we extract a number of features inspired by the eye tracking studies:~the x and y coordinates of the centre point and the radius of the AOI, time spent in the AOI (estimated using length of the uttered word), time to approach the AOI (estimated using word start time), number of visits to the AOI, transition time (similar to saccade time) and total length of the pauses (silence) made up to the current uttered word. Table~\ref{tab:features} summarises the extracted features. In addition to the AOI features, two other feature types were extracted, Age of Acquisition (AoA, the average age we normally learn a word for the first time \cite{forbes2005age}) and word vector features GloVe \cite{pennington2014glove}. We include the mean and standard deviation of AoA. The predefined GloVe word vectors are in 300 dimensions, however, using the Principal Component Analysis (PCA), the dimensions of the vectors were reduced to 7. Table~ref{tab:features} summarises all of the features. For each feature in the table the mean, standard deviation, minimum and maximum were calculated. 

\begin{table}
\caption{\label{tab:features}Extracted features: AOI: Area of interest. AoA: Age of Acquisition. WV: Word vectors}
\centering{
\begin{tabular}{p{2.4cm}|p{0.5cm}|p{3.6cm}}
\specialrule{0.8pt}{0.8pt}{0.8pt} 
\multicolumn{1}{c|}{\textbf{Feature}} &  
\multicolumn{1}{c|}{\textbf{Type}} &  
\multicolumn{1}{c}{\textbf{Description}}\\ 
\specialrule{0.8pt}{0.8pt}{0.8pt} 
$x\_coordinate$ & AOI & x coordinate of the AOI\\ 
\hline 
$y\_coordinate$ & AOI & y coordinate of the AOI\\ 
\hline
$radius$ & AOI & radius of AOI circle\\ 
\hline 
$time\_spent$ & AOI & time spent on the AOI\\ 
\hline 
$time\_to\_approach$ & AOI & time to approach the AOI\\ 
\hline 
$number\_of\_visits$ & AOI & number of visits to the AOI\\ 
\hline
$transition\_time$ & AOI & transition time from previous to current AOI\\ 
\hline
$pause\_length$ & AOI & the total pause length up to the current AOI\\ 
\hline
$mean\_aoa$ & AoA & mean of the AoA for a word\\ 
\hline
$std\_aoa$ & AoA & standard deviation of the AoA for a word\\ 
\hline
$wv_{1},...,wv_{7}$ & WV & GloVe word vector representation\\ 
\specialrule{0.8pt}{0.8pt}{0.8pt}
\end{tabular}
} 
\end{table}

\vspace{-2mm}
\section{Results}  
\label{sec:results}
The Kaldi ASR~\cite{Povey_ASRU2011} toolkit was used for training the ASRs. We followed the GMM/HMM based (SAT training) and the hybrid GMM/DNN (TDDN-LSTM) recipes. For the language models, we trained in-domain 3/4 grams with the KN/Turing smoothing. The HMM/GMM SAT training for the Dem dataset resulted in an average 64.2\% WER, while using the LSTM-TDNN recipe the average \textbf{41.6\%} WER  was achieved. For the IVA dataset a 33.8\% WER was gained by the DNN based ASR (50.5\% using SAT). For classifications, a Logistic Regression (LR) classifier was used using a 10-fold speaker independent cross-validation approach (none of the recordings of the speakers in the testing set are seen in the training set).

\subsection{Classification results using the forced alignments}
Table \ref{tab:LR_results_FA} shows the accuracy (Ac), recall (Rc), precision (Pr) and F1-score of the classifiers when using the forced alignment timing information. For the Dem dataset using only WV, only AoA and only AOI, 71\%, 62\% and 76.5\% F1-scores were achieved respectively; on their own the picture prompt based AOI features were the most discriminative for the AD/HC classification. As we added other feature types, the performance of the classifier improved, and the best performance was achieved by combining all features, resulting in an F1-score around 80\%. For the \iva{} dataset, using all the features, a 60.2\% F1-score was gained. 
Training the classifiers on a combination of the two datasets (the last two rows) not only did not improve the performance, but somewhat decreased the performance especially for IVA dataset.

\begin{table} 
\caption{\label{tab:LR_results_FA}LR classification results using forced alignment. All:~all features; Mix:~Dem+IVA; Ac:~Accuracy; Rc:~Recall; Pr:~Precision; F1:~F1-measure;}
\centering{
\begin{tabular}{p{1.4cm} | p{1.2cm}  | p{0.7cm} |p{0.7cm}  |p{0.7cm} |p{0.7cm} }
\specialrule{0.8pt}{0.8pt}{0.8pt} 
\multicolumn{1}{c|}{\textbf{Train/Test}} & 
\multicolumn{1}{c|}{\textbf{Feature}} &  
\multicolumn{1}{c|}{\textbf{Ac\%}}& 
\multicolumn{1}{c|}{\textbf{Rc\%}}& 
\multicolumn{1}{c|}{\textbf{Pr\%}}& 
\multicolumn{1}{c}{\textbf{F1\%}}\\
\specialrule{0.8pt}{0.8pt}{0.8pt}  
Dem/Dem & WV & 71.9 & 72.7 & 71.4 & 71.0\\  

\hline 
Dem/Dem & AoA & 63.2 & 63.2 & 62.8 & 62.0 \\  

\hline 
Dem/Dem & AOI & 77.3 & 78.1 & 76.8 & 76.5 \\ 

\hline 
Dem/Dem & AOI+AoA & 78.8 & 79.2 & 78.7 & 77.9 \\ 

\hline
Dem/Dem & AOI+WV & 79.2 & 79.7 & 78.7& 78.3\\ 
\hline
\hline
Dem/Dem& All & \textbf{80.8} & \textbf{81.1}  & \textbf{80.1} & \textbf{79.9}  \\  
 

\hline
IVA/IVA& All & 65.8 & 62.5( & 63.3 & 60.2 \\
 
\hline
Mix/Dem& All & 79.6 & 80.2  & 79.1  & 78.8  \\
\hline
Mix/IVA& All & 65.8 & 62.5  & 52.5  & 54.3   \\
\specialrule{0.8pt}{0.8pt}{0.8pt} 
\end{tabular}
} 
\end{table}

\subsection{Classification results using the ASR outputs}
Table \ref{tab:LR_results_DEC} shows the classification performance using all features extracted from the outputs of the ASRs (the words, the start time and the length of the words). As can be expected when comparing to Table \ref{tab:LR_results_FA}, the performance of the classifiers decreases, but not considerably. The F1-measure for Dem/Dem decreased around 8\% from around 80\% to 72\% , for IVA/IVA around 3\% from 60\% to 57\%. Mixing datasets, however, improved the performance for IVA from 54\% to around 64\%, but deteriorated for Dem (from 79\% to 71\%).

\begin{table} 
\caption{\label{tab:LR_results_DEC}LR classification results using the outputs of the ASRs and all features.~Mix:~Dem+IVA; Ac:~Accuracy; Rc:~Recall; Pr:~Precision; F1:~F1-measure;}
\centering{
\begin{tabular}{p{1.7cm} | p{0.7cm} | p{0.7cm}  |p{0.7cm}  |p{0.7cm} }
\specialrule{0.8pt}{0.8pt}{0.8pt}  
\multicolumn{1}{c|}{\textbf{Train/Test}} & 
\multicolumn{1}{c|}{\textbf{Ac\%}}& 
\multicolumn{1}{c|}{\textbf{Rc\%}}& 
\multicolumn{1}{c|}{\textbf{Pr\%}}& 
\multicolumn{1}{c}{\textbf{F1\%}}\\
\specialrule{0.8pt}{0.8pt}{0.8pt}  

Dem/Dem & \textbf{73.1} & \textbf{73.4} & \textbf{72.4} & \textbf{72.2} \\ 
\hline  

\mbox{IVA/IVA}  & 62.5 & 62.5 &  60.0 & 56.8  \\  
\hline

\mbox{Mix/Dem}  & 72.3 & 72.0  & 71.4  & 70.9  \\  
 
\hline
\mbox{Mix/IVA}  & 70.0 & 70.0 &  65.0 & 63.5  \\  
 
\specialrule{0.8pt}{0.8pt}{0.8pt} 
\end{tabular}
} 
\end{table}

\vspace{-2mm}
\section{Conclusions}
\label{sec:conclusions}
This paper presented an approach to aligning verbal picture descriptions with their picture prompt in order to estimate the attention and elicitation paths on the picture. Features, inspired by studies on the use of eye tracking devices for detecting cognitive decline in people with Alzheimer's disease, are then extracted. These features are then used to train a classifier achieving an 80\% F1-score when using information produced from the forced alignment of ASR and 72\% F1-score when we directly use the output of the ASR. In future work, we plan to investigate  the use of additional features to improve the classification and apply a DNN based sequence classifier.
\vspace{-2mm}
\section{Acknowledgements} 
This work is supported by the MRC Confidence in Concept Scheme and the European Union{\textquotesingle}s H2020 Marie Skłodowska-Curie programme TAPAS (Training Network for PAthological Speech processing; Grant Agreement No. 766287).  
\newpage

\bibliographystyle{unsrt}  






\bibliography{references}

\end{document}